\documentclass[runningheads]{llncs}
\usepackage[T1]{fontenc}
\usepackage{amsmath}
\usepackage{graphicx}
\usepackage{amssymb}

\begin{document}
\title{PointVoxelFormer - Reviving point cloud networks for 3D medical imaging}
%\author{No Author Given\inst{1}\orcidID{0000-1111-2222-3333}}
\author{Mattias Paul Heinrich\orcidID{0000-0002-7489-1972}}
%
%\authorrunning{N.A. Given}
\authorrunning{M.P. Heinrich}
%\institute{Princeton University, Princeton NJ 08544, USA
%\email{lncs@springer.com}}
\institute{University of L\"{u}beck, Germany
\email{mattias.heinrich@uni-luebeck.de}}

\maketitle              % typeset the header of the contribution
\begin{abstract}
Point clouds are a very efficient way to represent volumetric data in medical imaging. First, they do not occupy resources for empty spaces and therefore can avoid trade-offs between resolution and field-of-view for voxel-based 3D convolutional networks (CNNs) - leading to smaller and robust models. Second, they provide a modality agnostic representation of anatomical surfaces and shapes to avoid domain gaps for generic geometric models. Third, they remove identifiable patient-specific information and may increase privacy preservation when publicly sharing data.
Despite their benefits, point clouds are still underexplored in medical imaging compared to volumetric 3D CNNs and vision transformers. To date both datasets and stringent studies on comparative strengths and weaknesses of methodological choices are missing. Interactions and information exchange of spatially close points - e.g. through k-nearest neighbour graphs in edge convolutions or point transformations - within points clouds are crucial for learning geometrically meaningful features but may incur computational bottlenecks. This work presents a hybrid approach that combines point-wise operations with intermediate differentiable rasterisation and dense localised CNNs. For deformable point cloud registration, we devise an early fusion scheme for coordinate features that joins both clouds within a common reference frame and is coupled with an inverse consistent, two-step alignment architecture.  
Our extensive experiments on three different datasets for segmentation and registration demonstrate that our method, PointVoxelFormer, enables very compact models that excel with threefold speed-ups, fivefold memory reduction and over 30\% registration error reduction against edge convolutions and other state-of-the-art models in geometric deep learning.
\keywords{Point clouds \and Graph networks \and Splatting}
\end{abstract}

\section{Introduction}

Medical image analysis has so far been predominantly dealt with using dense voxel grid representations and machine learning approaches that are tailored towards such structured data. Convolutional neural networks (CNNs) and many types of vision transformers (ViTs) \cite{dosovitskiy2020image} - including the patch embedding, shifted window self-attention \cite{liu2021swin} or convolutional patch merging \cite{xie2021segformer} - all rely on a fixed and repetitive neighbourhood of image voxels (or pixels) to infer filter weights for learning visual features. Limitations of those dense 3D image analysis methods are high computational demand, the requirement of large labelled training datasets \cite{Tang_2022_CVPR} and a general tendency to fail to generalise in the presence of domain shifts due to overfitting.

Sparse representations in the form of unordered point clouds have tremendous benefits in terms of their efficient way of storing and processing only relevant locations that e.g. coincide with surfaces. Hence a volumetric CT scan with several million voxels can oftentimes equivalently be represented by few thousands 3D coordinates. Another advantage of point clouds is a modality agnostic representation, which can be used to learn geometric models from more diverse datasets given the input scans are converted into contours or surfaces. Dynamic graph convolutional networks (DGCNN) learn edge features by combining information for every pair of nodes connected in an underlying graph \cite{wang2019dynamic,bronstein2017geometric}. Point cloud approaches can also in principle benefit from new learning paradigms such as self-attention by treating each node neighbourhood as a set, which acts as local context (sequence). This has e.g. been demonstrated in the popular Point Transformer architecture \cite{zhao2021point}.

Geometric learning is, nevertheless, limited by a number of shortcomings. First, the neighbourhood definition usually relies on k-nearest neighbour (kNN) graphs that are very expensive to compute for larger clouds and will yield another hyperparameter $k$ that is hard to choose as it may be affected by memory limitation. It  also influences the achievable accuracy of a model. Second, the intermediate operation of such neighbourhood interactions are extremely inefficient in terms of memory usage. \cite{liu2019point} pointed out that irregular sparse memory access - inherent to all kNN based geometric approaches - results in a majority of run times, sometimes dwarfing the effort for the actual mathematical computations. This may lead to unnecessarily small point clouds that in turn yield underwhelming performance in comparison to dense CNNs or ViTs. On the other end, highly efficient simpler architectures such as the original PointNet \cite{qi2017pointnet} that solely focus on local pointwise multi-layer perceptrons (MLPs) fail to capture enough spatial context to perform competitively. Interestingly, recent work that analyses the power of vision transformer highlighted that local MLPs together with residual connections and any form of intermediate \textit{token mixing} are sufficient to reach state-of-the-art performance. First, \cite{dong2021attention} show that transformers without MLP or residual connections converge exponentially slower - hence attention is not really all you need. Second, \cite{yu2022metaformer} introduce the MetaFormer framework that outperforms ResNets (CNNs) and certain ViTs despite removing self-attention and replacing it with a simple (untrained) spatial pooling operation, while keeping the residual connections and channel MLPs. 

\subsection{Related Work:}
A number of architectures have been proposed over the last years to address the classification, segmentation and registration of point clouds, predominantly in the domain of general 3D vision but also with some success in medical imaging. The seminal \textbf{PointNet}\cite{qi2017pointnet} addresses the challenge of the permutation invariant and unordered structures of point clouds by solely relying on local feature learning with pointwise MLPs. In addition, it introduced two learnable rotation matrices that globally align spatial input and intermediate features respectively. It also uses one global pooling layer to gather contextual information for segmentation tasks or similar. Later the PointNet++ \cite{qi2017pointnet++} was proposed to extend the approach by including a pooling operation over local neighbourhoods. While improving upon the original PointNet its irregular memory access introduces an inefficiency of 80-90\% that quadruples runtime and memory requirements according to \cite{liu2019point}.

\textbf{Diffusion convolutional networks} introduce a neighbourhood using an isotropic smoothing on a graph that is defined by connecting each point (node) with its $k$ nearest neighbours \cite{atwood2016diffusion}. The filtering can be either performed by computing the graph Laplacian based on the (weighted) kNN-graph or a simple averaging over the gathered feature values from neighbouring nodes. Extensions include multiple smoothing kernels and increase the depth of the networks. They led to some improvements for human pose segmentation \cite{hansen2018multi}, but still fail to learn more complex edge features.

 \textbf{Dynamic Graph neural networks} (DGCNN) introduced edge convolutions \cite{wang2019dynamic} that are realised by performing a concatenation of features from all $k$ neighbouring nodes before feeding them into a shared MLP. Subsequently a local max pooling is used to reduce the processed information and obtain a new feature vector for each node. While this improves learning capacities it also expands memory use by a factor of $2k$, which means DGCNN are usually only applied to relatively small point clouds. \textbf{Graph Attention networks (GATs)}\cite{velivckovic2018graph} extend upon DGCNNs by extending the feature concatenation by a local computation of attention weights - potentially employing multi-heads, similar to transformers but with the difference that no different weighting matrices are used for key, query and value. GATs demonstrated good performance on brain parcellation (segmentation) using surface meshes.
 
GATs have subsequently been extended to {Point Transformers} \cite{zhao2021point}, which employ different key-query pairs of the current node and its neighbours to weigh the values across edges. They still suffer from inefficient memory access and costly re-computation of kNN-graphs for large point clouds, but improve on this in a practical manner by using a multiscale architecture with specialised down- and upsampling operations. The newest version of Point Transformer V3 \cite{wu2024point} linearises the irregular neighbourhood with space-filling Hilbert curves to improve memory efficiency. They thereby convert unstructured sparse data into structured 1D data akin to sequential language tokens to preserve the spatial proximity for the attention mechanism. Nevertheless the 1D arrangement is not trivial and requires dataset specific design choices for reordering, patch grouping with either shifted or shuffled orders to enable multiple different 1D Hilbert curves that jointly capture the true neighbourhood in all three dimensions. PointMamba \cite{liang2024pointmamba} uses both space-filling Hilbert and Trans-Hilbert curves and kNN for patches to extract token embedding for non-hierarchical state-space model learning (Mamba) and thereby improve over previous transformer architectures on point cloud segmentation. 

An orthogonal research direction was first presented in Vox2Vox PoseNet \cite{moon2018v2v} that first \textbf{rasterises} any high-resolution input point cloud into a low-resolution 3D occupancy map and subsequently applies a standard 3D CNN (e.g. UNet) to produce (keypoint) predictions. The method does not require a differentiable rasterisation and employs no interpolation, hence the accuracy is limited to the voxel-grid resolution. For 3D hand and human keypoint regression using heatmaps this coarse estimation is usually sufficient since the objects can be well captured within a cubic region of interest.

\begin{figure}[]
    \centering
    \includegraphics[width=0.9\linewidth]{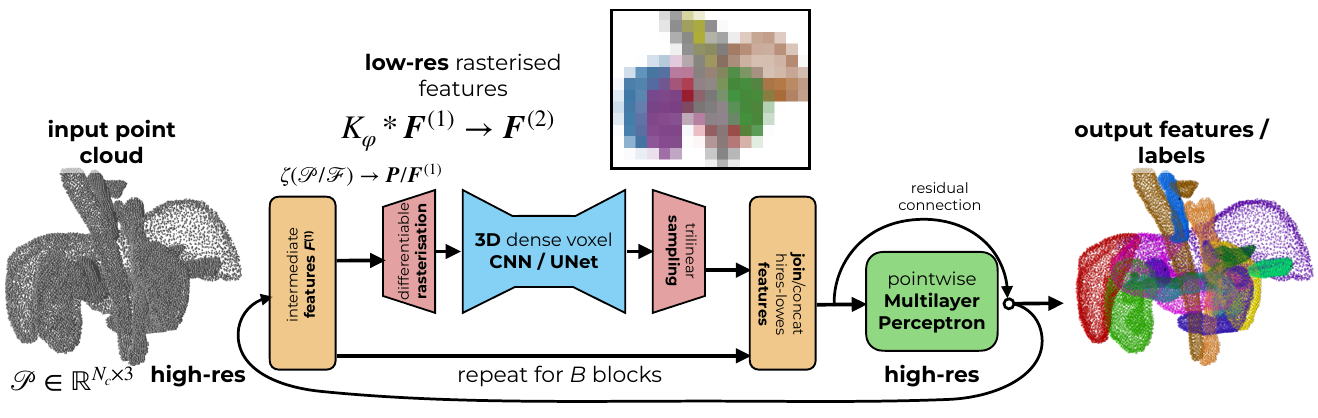}
    \caption{Overview of our proposed geometric learning framework \textbf{PointVoxelFormer} that alternates between point-wise MLPs on sparse high-resolution point cloud features and differentiable rasterisation coupled with dense shallow 3D CNNs to efficiently incorporate neighbour information. }
    \label{fig:concept-label}
\end{figure}

%Vox2Vox PoseNet, DGCNN, PointTransformer (+V2/3, Mamba), MetaFormer, 

\subsection{Motivation and Contributions}
Our hypothesis is therefore as follows: an optimal trade-off for point architectures that excel at 3D medical image analysis can be obtained by combining compute-efficient local (point-wise) MLP operators on relatively large-scale point clouds with alternating memory-efficient local pooling or convolution operations. To realise an efficient pooling operator on sparse irregular point clouds we remind ourselves of the concept of general blurring filters on high-dimensional data, e.g. the joint bilateral filter or permutohedral lattice, that combine splatting onto a regular simplex - in our case a low-resolution voxel grid - with a simple smoothing and subsequent sampling (slicing) step. As pointed out by Kopf et al. \cite{kopf2007joint} the paradigm works for joint upsampling technique, where a low-resolution input determines splatting positions, whereas the slicing is performed using a high-resolution reference image. This leads to an interpolation that, e.g. respects edges in the high-resolution data. A similar conceptional idea was proposed by \cite{liu2019point} in PointVoxelCNN, which voxelise intermediate point cloud features into a coarse regular 3D grid and subsequently fuses it with high-resolution point features. Using a 3D CNN on the coarse voxelised point clouds enables a learnable pooling of neighbourhood information. However, our approach extends this idea by the following important aspects:
\begin{enumerate}
    \item First, we incorporate the residual connections and transformer-like MLPs that are applied in an alternating rather than parallel fashion.
    \item  Second, contrary to the nearest neighbour rasterisation operation in PointVoxelCNN we propose to perform a differentiable rasterisation that enables sub-voxel accuracy and potential gradients with respect to the point coordinates for improved convergence.
    \item Third, we substantially extend the capabilities by flexibly varying the rasterisation grid size within the network - and extend the application from point cloud segmentation to deformable 3D registration.
\end{enumerate}

%reaching a 30 fold higher computational efficiency (3-4 times speed up) and five-fold lower memory requirements than graph-based geometric networks while at the same time substantially improving accuracy for segmentation and registration (DGCNN incurs an around 50% higher reg. error)

\section{Methods}

In this work we consider two complementary tasks for 3D medical image analysis, namely semantic point cloud segmentation and deformable point cloud registration. For the first task, we use a sparse point cloud $\mathcal{P}\in\mathbb{R}^{N_p\times 3}$  with $N_p$ 3D coordinates $p$ as input. We aim to learn a neural network $f$ with parameters $\phi$ to predict the class probabilities $\boldsymbol{p}_c = f(\mathcal{P},\phi)$ of each point to belong to a certain label $l\in\mathcal{C}_l$. For point cloud registration we specify a source cloud $\mathcal{S}\in\mathbb{R}^{N_s\times 3}$ and target cloud $\mathcal{T}\in\mathbb{R}^{N_t\times 3}$ as input and want to find a function that predicts $\boldsymbol{\psi} = f(\mathcal{S},\mathcal{T},\phi)$ a relative displacement field  $\boldsymbol{\psi}\in\mathbb{R}^{N_s\times 3}$ that geometrically aligns the source towards the target given a set of trainable parameters $\phi$. Both tasks will require the use of intermediate feature representations $\mathcal{F}\in\mathbb{R}^{N\times c}$ with a certain number of channels $c$ for the singular or joint point cloud through a number of operators.  

\begin{figure}[b]
    \centering
    \includegraphics[width=0.9\linewidth]{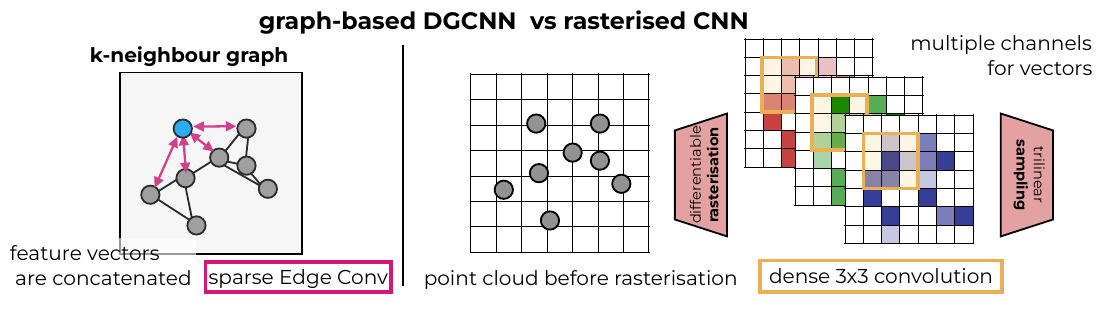}
    \caption{Comparison between previous work on dynamic graph CNNs (DGCNN) which need to compute kNN and gather features irregularly and our approach that first rasterises the feature maps to enable more efficient neighbourhood interactions using dense CNNs. Subsequently the results are sampled at high-resolution point coordinates using trilinear interpolation }
    \label{fig:raster-label}
\end{figure}

Assume that we are given a dense volumetric representation of the sparse $\mathcal{F}$ as $\boldsymbol{F}=\zeta(\mathcal{F})$. Now, convolutional neural networks on dense regular grids may learn filter kernels $K_\varphi$ with a fixed set of offsets - usually $3\times 3\times 3$,  yielding a dense operation $\boldsymbol{F}^{(2)}=K_\varphi*\boldsymbol{F}^{(1)}$ this is not straightforward for the unordered fashion in which point clouds are represented. PointNets \cite{qi2017pointnet} restrict the learnable operator to local MLPs or so-called $1\times 1$ convolutions - which severely restricts their spatial reasoning abilities. 

\textbf{Graph Convolutions}: Therefore, it is common to compute a filter kernel indirectly by first defining a spatial neighbourhood with the points $p_i\in\mathcal{N}(p_j)$ that are close to a specific coordinate $p_j$. The neighbourhood can be represented as a graph that provides the indices of the $k$ closest points based on Euclidean distances for every point. Next, the feature vectors $(x_i,x_j)$ of these points are extracted using a gather operation and, e.g. concatenated to the current features yielding a tensor of shape $N\times k\times 2\cdot c$. For \textit{Edge Convolutions} \cite{wang2019dynamic} a trainable nonlinear function $\varphi$ is subsequently applied that may comprise multiple layers, before aggregating the results over the dimension of the neighbours with a symmetric operator - usually max-pooling - denoted here as $\bigoplus$ - yielding $y_j=\underset{i\in\mathcal{N}_j}\bigoplus \varphi(x_i||x_j)$. 
One common extensions is the use of transformer-like architectures, i.e. creating multiple mappings of the input features called query, key and value for which a scaled dot-product attention is computed over the neighbourhood $\mathcal{N}$. The size $k$ of the graph can be seen as the sequence length. Another option are positional encodings based on the relative offsets of the original spatial coordinates compared to their considered neighbours. These are added - e.g. in PointTransformers - to the intermediate feature representation to strengthen convolution-like learning of local patterns in point clouds. Furthermore, graph convolutions or point transformers can be combined with local (PointNet-like) MLPs to increase the capacity of the network.

\textbf{Rasterisation: }The aforementioned approaches have a large intermediate increase in memory use and highly inefficient sparse gather access operations in common and yield very complex, hard-to-train networks. They provide a competitive baseline for certain tasks, but we aim for a more efficient solution that generalises better across 3D medical image analysis. An alternative solution that was first popularised in the Vox2Vox PoseNet and later extended within PointVoxelCNNs (PVCNNs) is to rasterise the high-resolution point cloud into a lower resolution grid $\boldsymbol{\hat{F}}=\hat{\zeta}(\mathcal{F})$ - the $\hat{ }$ symbol denoting an operation that reduces the quality of its inputs. Vox2Vox directly uses a 3D U-Net on the rasterised input volume, whereas PVCNNs repeats the use a nearest neighbour operator and averaging of all intermediate features from points that fall into a certain voxel. Subsequently a small CNN with usually two layers is applied to these voxelised features. PVCNNs have the advantage of maintaining the local high-resolution point information, process it using point-wise MLPs and fuse (implemented as addition) their result after each block with the \textit{devoxelised}, i.e. spatially sampled, features obtained from the low-resolution grid. Compared to DGCNNs the computational complexity is substantially lowered. Yet by rounding to nearest neighbours the approach is sensitive to the grid resolution.

Another approach to point cloud rasterisation called DiVRoC was recently presented in \cite{heinrich2023chasing} for the computation of an alternative loss that provides higher efficiency and better convergence than the widely used Chamfer distance for large scale point cloud registration. Different to PVCNNs the rasterised volume is constructed using the reverse operation of a trilinear interpolation. That means the information of each coordinate is distributed across all eight immediate (integer) neighbours according to their respective interpolation weights. The implementation makes use of the Jacobian of the \texttt{gridsample} operator in pytorch and yields a highly efficient trilinear splatting operations. Yet in this work it is only used for points with unit weight and not explored on intermediate features.

\subsection{PointVoxelFormer}Our approach combines several of the previously discussed concepts into a more generic hybrid point cloud analysis framework. A key aspects is to extend the splatting operation proposed in \cite{heinrich2023chasing} to intermediate multi-channel point features. Let us start with volumetric data that resides on a regular cartesian grid with integer coordinates $\boldsymbol{q}=(q_x,q_y,q_z)\in\mathbb{N}^3$ and associated voxel intensities $\boldsymbol{x}\in\mathbb{R}^{n_x\times n_y\times n_z \times c}$, where $c$ denotes the channels. We follow the notation of \cite{dai2017deformable} to define a trilinear interpolant $G(\boldsymbol{q},\boldsymbol{p}) = g(q_x,p_x)\cdot g(q_y,p_y)\cdot g(q_z,p_z), $ with $g$ being defined as $g(q_x,p_x)=\operatorname{max}(0,1-|q_x-p_x|)$, which can be used to perform a weighted average over the 8 neighbouring (integer) grid points closest to our floating point coordinate $\boldsymbol{p}$: $\boldsymbol{y}(\boldsymbol{x}) = \sum_{\boldsymbol{q}}G(\boldsymbol{q},\boldsymbol{p})\cdot\boldsymbol{x}(\boldsymbol{q}).
$. Hence as detailed in \cite{heinrich2023chasing} for the case of sparse 3D point clouds $\mathcal{P}$, the order of this equation has to be reversed to $\boldsymbol{x}(\boldsymbol{q}) = \sum_{\boldsymbol{p}}G(\boldsymbol{p},\boldsymbol{q})\cdot\boldsymbol{y}(\boldsymbol{p})$ resulting in a trilinear splatting operation $\boldsymbol{P}=\boldsymbol{x}(\boldsymbol{P})=\zeta(\mathcal{P})$. %When used in isolation this may lead to visually undesirable effects, e.g. holes in the output and multiple intensities that are accumulated onto the same location.

We integrate this new method into a hybrid multi-layer point-voxel architecture and use a concatenation to fuse features. In contrast to PVCNNs but more similar to MetaFormers, we apply the point-wise MLP not in parallel to the rasterised 3D CNN but alternate between both operations to avoid a competition between the two branches. In addition, we make another practical adjustments for the most challenging deformable registration task by employing a sequence of incrementally increased grid sizes for the voxelisation.

\subsubsection{Deformable registration with generic point cloud architectures} Deep learning based image registration most often uses a concept for joining the information of fixed and moving images by concatenating their input channels and feeding them into a U-Net for displacement prediction. This is not directly possible for point clouds and many approaches so far have resorted to complex hand-crafted methods for combining the information of both inputs. This includes e.g. cross-attention, optimal transport \cite{shen2021accurate} and correlation layers \cite{wu2020pointpwc}. Here we present a surprisingly simple new solution that expands the input features channels with three leading zeros for each of the source points and three trailing zeros for target cloud points before concatenating those in the set dimensions leading to a tensor of shape $(N_s+N_t)\times 6$.

\subsubsection{Multi-step, regularisation and inverse consistency:} To ensure regularity while still capturing large deformations, we consider a two-step inverse consistent approach by construction following \cite{greer2023inverse}. To achieve this, we first interpolated the sparse motion fields and implicitly regularise it using a dense B-spline transformation. Next we
estimate a symmetric field $\varphi^*=\exp(f_{\mathcal{TS}}-f_{\mathcal{ST}})$, where $\exp$ represents the exponentiation step of the scaling-and-squaring approach. When estimating large deformations a single network is usually unable to capture the complex transformation. Hence, a two-step consistent approach with two networks $f^{(1)}$ and $f^{(2)}$ is used:
\begin{equation}
\varphi^{(2)}=\exp(\frac{f^{(1)}_{\mathcal{TS}}-f^{(1)}_{\mathcal{ST}}}{2})\circ\exp(f^{(2)}_{\mathcal{\bar{T}\bar{S}}}-f^{(2)}_{\mathcal{\bar{S}\bar{T}}})\circ\exp(\frac{f^{(1)}_{\mathcal{TS}}-f^{(1)}_{\mathcal{ST}}}{2}).
\label{eq:twostep}
\end{equation}
Here, two mid-way warps are defined as $\bar{\boldsymbol{S}}=\zeta(\mathcal{S}\circ\exp(\frac{f^{(1)}_{\mathcal{ST}}-f^{(1)}_{\mathcal{TS}}}{2}))$, similar to the ANTs SyN approach \cite{avants2008symmetric}. More details on the implementation of each step can be found within our open source code available in anonymous form at \url{https://anonymous.4open.science/r/PointVoxelFormer-0E02}.

\section{Experiments}
To comprehensively evaluate the benefits of rasterisation within 3D point cloud analysis we create and make publicly available two new derived datasets for medical point cloud segmentation and further use the public 3D deformable lung registration dataset PVT1010 \cite{shen2021accurate}.

\textbf{3D Ultrasound segmentation }based on the public SegThy dataset \cite{kronke2022tracked}. The first experiments are based on the excellent 3D ultrasound dataset for thyroid assessment that comprises scans from 186 patients and manual as well as automatic segmentations for thyroid, carotid arteries and jugular veins. Ultrasound examinations have a high dependency on the user’s experience and may require realtime guidance for non-experts. Hence, exploring 3D point cloud algorithms that can segment volumes with 50 frames per second and more with the potential to be implemented on the end-device - compared to several seconds on a high-end GPU for 3D dense voxel U-Nets - are of relevance. The original scans have dimensions of approximately 400$^3$ voxels with an isotropic resolution of 0.12~mm. We downsample the scans to 0.24~mm and extract an edge-map for each 2D slice individually using skimage’s Canny edge detector with $\sigma=5$ and stack them into a 3D volume again. From this we can extract roughly 300 thousand 3D points per volume that we reduce using furthest point sampling to 16'384 (2$^{14}$) coordinates. To obtain foreground and background labels we dilate the segmentation with a spherical structuring element of $9\times 9\times 9$ voxels before assigning the labels to the points. We have overall 32 annotated volumes - with multi-label manual expert segmentations - which we split into 24 training and 8 test cases. The background covers roughly 86\% of the points whereas the thyroid comprises 8\% and veins or arteries the remaining 6\% - resulting in a fairly unbalanced and thus challenging segmentation task.
\begin{figure}
    \centering
    \includegraphics[width=0.95\linewidth]{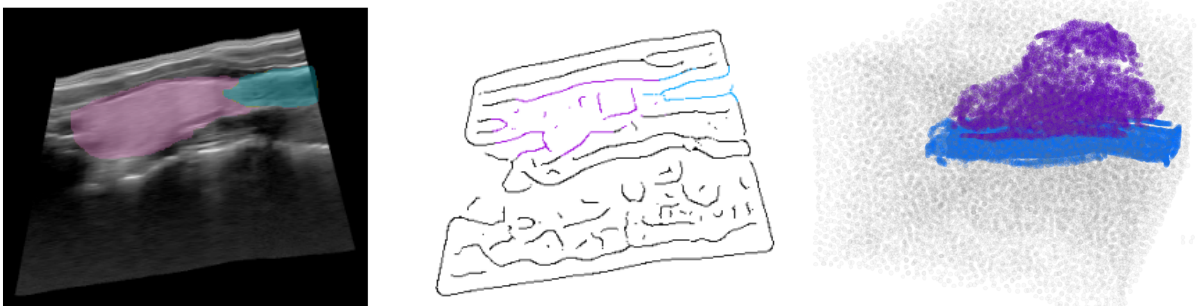}
    \caption{Automatic extraction of 3D point clouds with edges in background (gray/black) based on original ultrasound scans (left) using the Canny edge detector (middle), morphological processing and point sampling (right).  }
    \label{fig:us-label}
\end{figure}
\textbf{Abdominal organ labelling }based on the TotalSegmentator dataset \cite{wasserthal2023totalsegmentator}. To increase the complexity of the anatomical labels to smaller organs of more difficult shape, we create a second segmentation dataset by extracting 163 volumes from the TotalSegmentator CT data that comprise at least 12 of the 13 abdominal organs used in the popular \textit{Beyond the Cranial Vault} segmentation challenge. We sample point clouds with 16'384 (2$^{14}$) coordinates from surfaces that we extracted from the provided ground truth segmentation. This time we do not include any background resulting in approx. 40\% liver points and just 0.5\% (70-90 points) for adrenal glands. We divide the dataset into 99 training and 64 test volumes. Before the advent of deep learning this task, solved using deformable registration and multi-atlas fusion of 30 labelled scans, resulted in approx. 79\% Dice overlap \cite{heinrich2015multi} for unseen images. Nowadays with large-scale vision transformers and additional pre-training on hundreds of CTs yield over 90\% Dice \cite{Tang_2022_CVPR}.

\textbf{Pulmonary vessel tree registration} using 2020 3D point clouds extracted from paired inspiration and expiration scans of 1010 COPD patients \cite{shen2021accurate}, which is also openly available \footnote{\url{https://github.com/uncbiag/shapmagn/} PVT1010 readme.md}. The dataset stands out for the high resolution with on average over 80 thousands coordinates per cloud. These are reduced to $2\times$ 24'576 for each pair. Together with a set of 10 patients from DirLab-COPD \cite{castillo2013reference}, with their corresponding two point clouds each and 300 manual landmark pairs annotated by experts in the original CT scans, they offer one of the most challenging registration tasks available. The initial misalignment is over 20~mm and no ground truth correspondences are available for the training data. We hence explore an unsupervised training using a new point cloud density cost function proposed in \cite{heinrich2023chasing}. 

\subsection{Compared methods} We benchmark our proposed PointVoxelFormer against PoolFormer, DGCNN, Vox2Vox - here refered to as PointUNet - and PointTransformer. In addition we evaluated several ablations regarding the proposed changes in comparison to PointVoxelCNN (PVCNN). For the two segmentation tasks each model, except PointUNet comprises eight blocks with MLPs that intermediately expand the initial number of feature channels ($C=96$) by four-fold. The rasterisation volume is set to $32 \times 32\times 32$ voxels for our approach and double that for PointUNet. PointVoxelFormer uses shallow CNNs with two $3 \times 3\times 3$ kernels for each block, whereas PointUNet uses a MONAI UNet with three levels after a single rasterisation. For PointTransformer and DGCNN we explore a range of $k$-nearest neighbours starting from 8 up until 24 (but not all fit into 32 GByte of VRAM for the DGCNN). PointVoxelFormer and PointTransformer also use a version with 128 channels. Both kNN based approaches are implemented using always the same (high-resolution) level and we replace the approximate linear attention in PointTransformer with a full Flash Attention \cite{dao2022flashattention}, which is similarly fast. As extension to the original implementation of DGCNN in \cite{wang2019dynamic}, we add a trainable positional encoding to each block, which much improved the capabilities of this baseline. A further baseline called PoolFormer is included that does not apply a learnable token-mixer in between MLPs but simply applies a nonlinearity to all neighbouring feature vectors in the kNN-graph before average pooling. All models are trained for 3000 iterations (120 epochs) using Adam with a learning rate of $10^{-3}$ and cosine annealing with a batch size of 4. Affine augmentation is applied for the smaller ultrasound thyroid dataset to avoid overfitting.

For the registration task, we increase the batch size to 5 and train for 10'000 iterations with a learning rate of $5\times10^{-3}$, i.e. for 50 epochs - and continue training for another 50 epochs with halved learning rate for the best models. We leave out PointTransformer due to its relatively poor performance on both segmentation tasks. Since the four-fold channel expansion also yielded no benefits and unnecessarily increased memory usage we also dropped this and further reduced the number of blocks to four and the channels to $C=64$. Different to before, the methods that perform intermediate rasterisation (PointVoxelFormer and PointVoxelCNN) are now implemented to use 4 blocks with increasing grid dimensions with 24$^3$, 32$^3$, 48$^3$ and 64$^3$ voxels. 

\begin{table}[b]
    \centering
    \begin{tabular}{l|l l| l l}
    \hline
        \textbf{Method $\downarrow$} & Accuracy & Dice (thyroid) & Accuracy & Dice (abdomen)   \\
        \hline
         PoolFormer&  89.11\%&54.95$\pm$28.2\%&76.80$\pm$42.2\% &60.31$\pm$19.4\% \\
         PointUNet&  94.08\% & 79.27$\pm$12.5\% & 90.05\%6.4\% & 78.86$\pm$11.07\% \\
         PointTransformer& 91.07\% & 68.96$\pm$18.2\% &88.42$\pm$32.0\% &78.86$\pm$11.9\% \\
         EdgeConv / DGCNN&91.81\%&71.16$\pm$16.82\%&91.53$\pm$27.8\% &83.89$\pm$10.1\% \\
         PointVoxelCNN&  93.86\%&77.93$\pm$14.9\%&91.31$\pm$6.9\%&81.03$\pm$13.52\%\\
         \hline
         \textbf{PointVoxelFormer}&\textbf{95.01\%}&\textbf{83.04$\pm$11.6\%}&\textbf{93.84$\pm$24.0\% }&\textbf{86.49$\pm$10.7\%}  \\
         \hline
    \end{tabular}
    \caption{Our comparison demonstrates the advantages of the new hybrid PointVoxelFormer against both kNN-graph based methods such as EdgeConv (DGCNN) and PointTransformer as well as alternative rasterisation approaches such as the Point-UNet (Vox2Vox) and the original PointVoxelCNN.  }
    \label{tab:segmentation}
\end{table}
\section{Results and Discussion}
In Table \ref{tab:segmentation} the result of both segmentation tasks for the best version of each tested model is listed. It demonstrates the advantages of our proposed PointVoxelFormer against all benchmark methods. Notably, DGCNN outperform PointTransformer and are second best on the relatively clean abdominal organ labelling. The PointUNet, however, has a higher performance than DGCNN - only second to our PointVoxelFormer - for the ultrasound thyroid segmentation. The model comprises 1.17 million trainable parameters and can be run with 112 frames per second on a single GPU that means each 3D segmentation takes less than 9 milliseconds. Compared to the DGCNN our approach reduces the required graphics memory by five-fold.

\begin{figure}[t]
    \centering
    \includegraphics[width=1\linewidth]{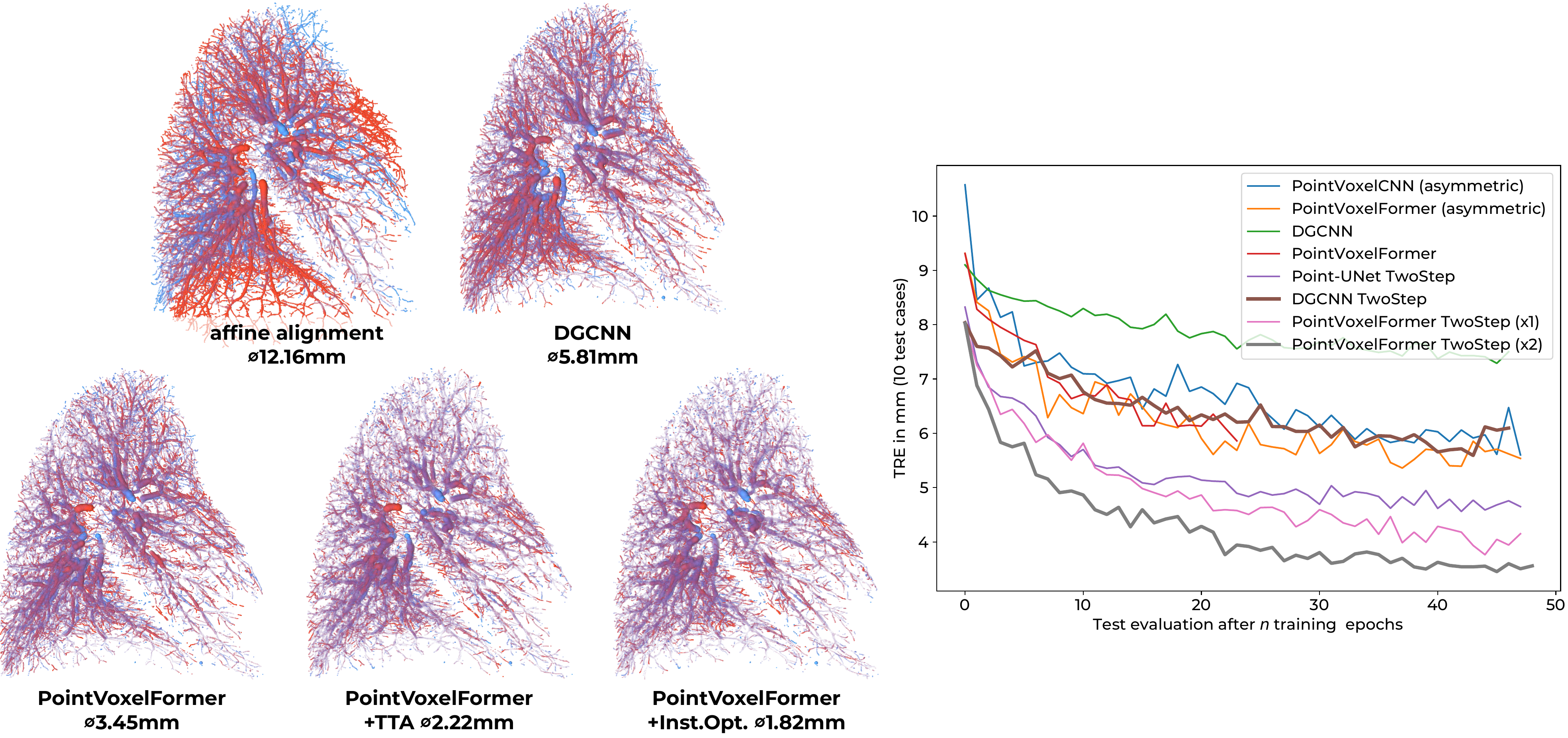}
    \caption{Left: Visual results for the PVT1010 registration task (Case \#4 of DirLab-COPD), showing overlay of inspiration (red) and expiration (blue) high-resolution point clouds. Purple indicates good alignment. The target registration errors are shown as average ($\varnothing$) over all ten cases in mm. Right: TRE results for the PVT1010 registration tasks evaluated at every epoch.}
    \label{fig:runtre-label}
\end{figure}
%PointTransformer
%train tensor(0.9752) tensor(0.9179)
%val tensor(0.9107) tensor(0.6896)
%tensor([0.9541, 0.6344, 0.6352, 0.5347])0.1825
%DIVROC96
%#val tensor(0.9478) tensor(0.8146)
%#w/o augmentation
%#val tensor(0.9202) tensor(0.6976)
%#DIVROC with 128 channels
%#val tensor(0.9501) tensor(0.8304)
%#tensor([0.9735, 0.7970, 0.8566, 0.6946]), 11.65
%#PointVoxelCNN
%Abdomen val tensor(0.9131) tensor(0.8103)
%tensor(0.7800) tensor(0.1352) tensor(0.9131) tensor(0.0696)

%#train tensor(0.9693) tensor(0.8993)
%#val tensor(0.9386) tensor(0.7793)
%#tensor([0.9678, 0.7591, 0.7859, 0.6044])14.9
%#EdgeConv with pos. encode k=8
%#train tensor(0.9675) tensor(0.8900)
%#val tensor(0.8998) tensor(0.6613)
%#EdgeConv with pos. encode k=16
%#train tensor(0.9808) tensor(0.9378)
%#val tensor(0.9181) tensor(0.7116)
%#tensor([0.9594, 0.6552, 0.6473, 0.5844])0.1682
%#SimplePoolBlock
%#train tensor(0.9665) tensor(0.8840)
%#val tensor(0.8911) tensor(0.5495)
%#tensor([0.9442, 0.5582, 0.3740, 0.3214]) 0.2821
%#UNET 64x64x64
%#train tensor(0.9880) tensor(0.9587)
%#val tensor(0.9408) tensor(0.7927)
%#tensor([0.9701, 0.7766, 0.7498, 0.6743])
%#w/o augmentation
%#train tensor(1.0000) tensor(0.9999)
%#val tensor(0.8899) tensor(0.5633)

\textbf{Ablation study for PVT1010: }We explore a number of variants for the considered point cloud network architectures and implementation choices for the registration model. As reference for these comparisons we chose the two-step symmetric registration model using the proposed PointVoxelFormer trained for 50 epochs that reaches 3.98~mm TRE. This is an improvement of 0.60~mm over the PointVoxelCNN and a gain of 1.09~mm compared to the PointUNet, which uses a single rasterisation with 64$^3$ voxels. The advantage compared the previous most widely employed DGCNN is even more striking with 2.04~mm. We could, however, not find a statistically significant difference between using nearest neighbour or trilinear rasterisation in this setting and therefore attribute the advantages of our method against PointVoxelCNN to the alternating rather than parallel use of point-wise MLPs. 

We also evaluated our model with only a single step (but still symmetric), which performs 1.22~mm worse than the two-step. An asymmetric version further deteriorates the TRE by 0.39~mm to 5.19~mm. All training runs are shown in Fig.~\ref{fig:runtre-label} over the 50-100 epochs along with visual examples of the aligned point clouds. The PointVoxelFormer registration model is smaller in size with 244 thousands parameters and runs at 125 fps per model, which results in approx. 30ms run time per registration for the symmetric two-step approach.

\begin{table}[t]
    \centering
    \begin{tabular}{c|c | c | c | c | c | c | c}
    \hline
        \textbf{ Model $\rightarrow$} & DGCNN & DGCNN & PointPWC & Spline & LDDMM  &  DGCNN & \textbf{PVFormer}\\
        \textbf{Mode $\downarrow$}& -CPD  & \multicolumn{2}{c|}{-DiVRoC \cite{heinrich2023chasing}} & \multicolumn{2}{c|}{-RobOT \cite{shen2021accurate}} & \multicolumn{2}{c}{\textbf{ours}}\\
        \hline
        %\textbf{Mode $\downarrow$}&{}&&&\\
        network-only & - & 7.35 & 5.96 & 5.72 & 5.48 & 5.81 & \textbf{3.45~mm} \\
        \hline
        +optimisation & 4.3 & 2.71 & 2.39 & 2.95 & 2.86 & 1.97 & \textbf{1.82~mm}  \\

\hline
    \end{tabular}
    \caption{Target registration error (in mm) comparison on PVT1010 compared to recent state-of-the-art demonstrating superior accuracy of the propose PointVoxelFormer. }
    \label{tab:registration}
\end{table}

\textbf{Best registration model with instance optimisation: }To finally obtain the best possible registration of this challenging PVT1010 task, we extend the training of both the DGCNN and our PointVoxelFormer model to 100 epochs, which yields an average TRE of 5.81~mm and 3.45~mm (ours) respectively. We compare these results without and with fine-tuning or instance optimisation to current state-of-the-art models that are also trained in an unsupervised fashion, namely the DGCNN and PointPWC models in \cite{heinrich2023chasing} and the robust optimal transport (RobOT) model also based on PointPWC with different fine-tuning options. When directly comparing the DGCNN models (our baseline) the results are similar to previous published methods: 5.5-7~mm. Yet, we obtain a substantial gain with our proposed PointVoxelFormer that outperforms previous unsupervised methods before instance optimisation by more than 2~mm. When applying test-time adaptation (TTA), i.e. training with the unsupervised loss on the test data, we improve our model to 2.22~mm. With a case-specific instance optimisation (35 iterations using a B-spline transformation model) we lower the TRE from 3.45 to 1.82~mm - setting a new state-of-the-art for the PVT1010 dataset - see Table \ref{tab:registration}. Note that this is also better than a recent supervised approach \cite{bigalke2023denoised} that simulates deformable lung motion and bridges the domain gap with a denoised mean teacher to train the PointPWC method and reaches a 2.31~mm TRE. In addition, we substantially outperform popular image-based deep learning registration on the same DirLab-COPD data - LapIRN \cite{mok2020large} and VoxelMorph \cite{balakrishnan2019voxelmorph}  with 4.76~mm and 7.98~mm respectively.  

%tensor([1.4835, 3.1311, 1.3947, 1.6895, 1.8214, 1.6183, 1.5859, 1.8759, 1.5216, 2.0433], grad_fn=<SelectBackward0>)
%tensor([1.4661, 3.1338, 1.3942, 1.7032, 1.7936, 1.6179, 1.5649, 1.8610, 1.5306,2.0407],

\section{Conclusion}
We demonstrated that point cloud networks, which are so far underexplored in medical imaging research, can offer a versatile and highly accurate solution for diverse segmentation and registration tasks. In particular for time-sensitive, modality-agnostic and privacy-concerning tasks they are an attractive alternative to dense voxel-based CNNs and vision transformers.

%\section*{Acknowledgments}
%Anonymised for peer review.
\bibliographystyle{splncs04}
\bibliography{samplebibliography}

\end{document}